# An Enhanced Randomly Initialized Convolutional Neural Network for Columnar Cactus Recognition in Unmanned Aerial Vehicle Imagery

Safa Ben Atitallah[a], Maha Driss[a,b], Wadii Boulila[a,b], Anis Koubaa[c], Nesrine Atitallah[d,e], Henda Ben Ghézala[a]

[a]*RIADI Laboratory, National School of Computer Sciences, University of Manouba, Tunisia*
[b]*IS Department, College of Computer Science and Engineering, Taibah University, Saudi Arabia*
[c]*Robotics and Internet-of-Things Laboratory, Prince Sultan University, Saudi Arabia*
[d]*Faculty of Computer Studies, Arab Open University, Saudi Arabia*
[e]*CES Laboratory, National Engineering School of Sfax, University of Sfax, Tunisia*

**Abstract**

Recently, Convolutional Neural Networks (CNNs) have made a great performance for remote sensing image classification. Plant recognition using CNNs is one of the active deep learning research topics due to its added-value in different related fields, especially environmental conservation and natural areas preservation. Automatic recognition of plants in protected areas helps in the surveillance process of these zones and ensures the sustainability of their ecosystems. In this work, we propose an Enhanced Randomly Initialized Convolutional Neural Network (ERI-CNN) for the recognition of columnar cactus, which is an endemic plant that exists in the Tehuacán-Cuicatlán Valley in southeastern Mexico. We used a public dataset created by a group of researchers that consists of more than 20000 remote sensing images. The experimental results confirm the effectiveness of the proposed model compared to other models reported in the literature like InceptionV3 and the modified LeNet-5 CNN. Our ERI-CNN provides 98% of accuracy, 97% of precision, 97% of recall, 97.5% as f1-score, and 0.056 loss.

## 1. Introduction

Internet of Things (IoT) refers to the connections between billions of objects over the internet [1]. These devices collect and share information based on standardized communication protocols, generating a huge amount of data with heterogeneous types. Investigating this data using Artificial Intelligence (AI) helps to provide efficient predictive analytics [2]. Recently, the capacities of AI have evolved and Machine Learning (ML), a subfield of AI, is presented. ML enables computer programs to learn from data and improve behaviors to make decisions [3]. Deep Learning (DL) is an advanced ML technique based on algorithms and functions of multilayer neural networks [4]. These techniques are employed to transform the raw data into actions and insights resulting in providing different forms of analytics: 1) descriptive, 2) predictive, and 3) prescriptive [5]. In this context, DL has extensively utilized in several case studies especially when working with complex and big amounts of data [6-8].

On the other hand, Remote Sensing (RS) has stood out as a breakthrough technology for large-scale earth observation and conservation. RS data applications cover a wide range of fields such as intelligent surveillance, urban planning, and sustainable forestry [9-12]. However, with the increased use of aerial and satellite sensors, the growing diversity and complexity of RS data have imposed multiple challenges [13] like data representation and interpretation. To overcome these challenges, DL techniques were performed on RS data and the experimentations that are carried out have demonstrated impressive achievements and results in multiple case studies [14][15]. Among the most used DL techniques, we cite the Convolutional Neural Networks (CNNs), which have shown competitive and remarkable performance in a wide variety of domains, especially for the image processing domain [16][17]. One of the case studies for which CNNs have been employed is the plant types and diseases' recognition. Indeed, developing autonomous surveillance systems for protected areas that record human activity and its effect on vegetation is in dire need nowadays in order to ensure the protection of biodiversity and the sustainable use of ecosystems. These systems could collect data using IoT devices and apply CNNs to perform the needed analytics.

Several substantial studies have been conducted about plant disease detection, plant family prediction, and plant recognition. For instance, Zhu et al. [18] propose a two-way attention method using CNN to predict the plant family labels and then determine the species labels. The proposed CNN is based on the Xception model with pre-trained weights. In [19], López-Jiménez et al. develop an approach for columnar cactus recognition. This approach relies on a modified version of LeNet-5 CNN. The used network was not deep, it consists of two convolutional and max-pooling layers, followed by a fully connected layer. In [20], Zheng et al. propose a recognition method based on the CNN model with different depths for the disease recognition of fruit trees. This method doesn't use RS images but relies on statistical and individual sample data of healthy and unhealthy leaves that are augmented to obtain an enhanced data set and improve the resulting accuracy. In [21], Nezami et al. have proposed 3D convolutional neural networks trained with hyperspectral and Red-Green-Blue channels to classify boreal tree species. The proposed 3D-CNN models were used to classify tree species in a test site in Finland. These models were performed only for tree species classification and not for individual tree detection. Based on the MobileNet and Faster R-CNN, Hu et al. [22] propose a method to detect the diseased pine trees from RS images. They used the MobileNet to reduce the interference of background information and trained the Faster R-CNN to identify the diseased pine trees. In terms of classification accuracy, the proposed R-CNN was compared with only traditional machine learning approaches, such as support vector machine and AdaBoost.

In the present work, we aim to detect the columnar cactus specifically the Neobuxbaumia tetetzo species in the Tehuacan-Cuicatlan Valley, which is a protected natural area located in southeastern Mexico. Columnar cactus is an endemic plant of the Valley which is classified as a mixed site in the World Heritage List by UNESCO in 2018. To recognize this type of plant from RS images, we propose an Enhanced Randomly Initialized Convolutional Neural Network (ERI-CNN). A randomized neural network is defined as a neural network with multi-layered architecture, where the connections between composing layers are untrained before the initialization [23]. To train and test the proposed model a huge dataset containing more than 21,000 RS images that are collected by an unmanned aerial vehicle from a height of 100 meters is used. The main contributions of this work can be summarized according to the following 3 points:
1. A novel randomly initialized CNN model for the columnar cactus recognition is designed;
2. The impact of data augmentation on model performance is evaluated;
3. A comparison between the developed network and other models in terms of effectiveness and performance is carried out.

The rest of the paper is organized as follows: Section 2 presents the proposed approach. Section 3 demonstrates the experimental implementation, the used dataset, and the obtained results. Section 4 reveals a comparison with other applied networks. Finally, Section 5 draws out concluding remarks and future works.

**2. Proposed Approach**

This section is devoted to provide an overview of CNNs, the proposed weight initialization method, and the architecture design of the proposed network.

*2.1. Convolutional Neural Networks*

CNNs have addressed their effectiveness to learn a high level of features from raw and complex data through a set of hidden layers. In the training phase, according to the input data, the input layer assigns randomly the initial weights which are passed to the next layer. The following layers perform filter operations and update the weights' values. In the end, the optimized weights are selected to be used to determine the final output. Fig. 1. presents the architecture of the CNN DL model.

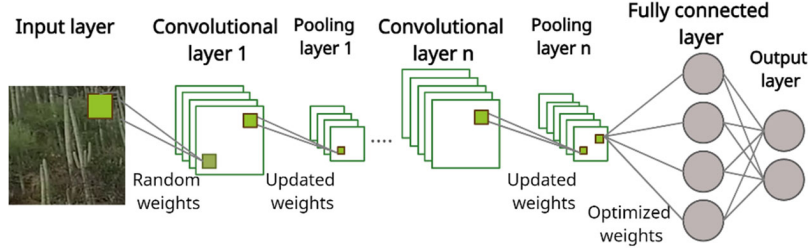

Fig. 1: Architecture of the CNN DL model

*2.2. Proposed Approach for Weight Initialization*

Tuning the right configuration parameters of a CNN model, such as the weights of the layers, ameliorates the network performance. Different types of weight initialization based on randomization have been proposed. Indeed, weight initializers are the regulating methods of the initial weights of a neural network [24]. A careful weight setting is required to get better results and avoid the gradient vanishing/exploding problem. Inspired by [25], we propose to improve the performance of CNNs by selecting the network weights based on an enhanced randomized method. Rather than starting by choosing weights, we begin by randomly selecting the slope angles for activation functions. The values of the selected slope angles are taken from the following interval:

$$\mathcal{I} = (-90°, -\alpha_{min}) \cup (\alpha_{min}, 90°) \tag{1}$$

where $\alpha_{min} \in [0°, 90°]$

After that, around the y-axis, the used activation function is randomly rotated. Therefore, the weights will be assigned and distributed over the input space in line with the input data distribution. Following this, the weights are calculated according to equation (2):

$$\omega_{i,j} = -4 \frac{w'_{i,j}}{w'_{i,0}}, \; j = 1,2,3,\dots,n \tag{2}$$

where $w'_{i,j}$ are the components of the normal vector n to the hyperplane that is the tangent to the activation function inflection points.

The rotation of the activation function is done randomly, computed by choosing the components of the normal vector in random and determining the $w'_0$ using equation (3):

$$w'_0 = (-1)^c \frac{\sqrt{(w'_1)^2 + \dots + (w'_n)^2}}{\tan \alpha} \tag{3}$$

where $c \sim U\{0, 1\}$

*2.3 Proposed Network Architecture*

To recognize the cactus plants, we propose an ERI-CNN that consists of six layers: an input layer, 4 hidden layers composed of convolutional and pooling layers, and a classification layer. The size of input images of the network is (32, 32, 3). For each convolutional layer, filters with a size of (2×2) and (3×3) are applied with padding. While every pooling layer implements a max-pooling window of a size of (2×2). In the following, "Conv2D" represents the convolution layer, "MaxPool2D" the pooling layer, and "FC" the fully connected layer. Table 1 illustrates the architecture of the proposed ERI-CNN.

Table 1: The architecture of the proposed ERI-CNN

| Layer | Description | Values |
| --- | --- | --- |
| Input layer | Images input layer | Input shape= (32,32,3) |
| Hidden layer 1 | 1$^{st}$ Conv2D<br>MaxPool2D | 16 feature maps (2×2)<br>Pool size= (2,2) |
| Hidden layer 2 | 2$^{nd}$ Conv2D<br>3$^{rd}$ Conv2D<br>MaxPool2D | 32 feature maps (2×2)<br>32 feature maps (2×2)<br>Pool size= (2,2) |
| Hidden layer 3 | 4$^{th}$ Conv2D<br>5$^{th}$ Conv2D<br>6$^{th}$ Conv2D<br>MaxPool2D | 64 feature maps (3×3)<br>64 feature maps (3×3)<br>64 feature maps (3×3)<br>Pool size= (2,2) |
| Hidden layer 4 | 7$^{th}$ Conv2D<br>8$^{th}$ Conv2D<br>9$^{th}$ Conv2D<br>MaxPool2D | 128 feature maps (3×3)<br>128 feature maps (3×3)<br>128 feature maps (3×3)<br>Pool size= (2,2) |
| Classification layer | 2 FC layers<br>Sigmoid | 128 units<br>1 unit |

## 3. Experiments

In this section, we detail the implementation of the proposed network and describe the used dataset. Then, we discuss the obtained results and examine the impact of data augmentation on the proposed model performance.

*3.1 Implementation Details*

In this study, the experiments are carried out using a PC with the following configuration properties: an x64-based processor; an Intel Core i7-8565U CPU @ 1.80GHz 1.99GHz; and a 16 GB RAM running on Windows 10 with NVIDIA GeForce MX. The networks are programmed with Jupyter notebook using python 3.7. We have employed both Keras library and TensorFlow backend. For faster computation, we used the Nvidia GeForce MX 250 GPU with CUDA and cuDNN libraries.

The developed ERI-CNN is trained over 100 epochs. For the model configuration, we adopted Adam optimizer with a learning rate value (1e-4) [26]. The binary cross-entropy function is used to measure the network performance on the training data, while the Rectified Linear unit (ReLu) is used for activation.

*3.2 Dataset Description*

The used dataset is Cactus Aerial Photos downloaded from Kaggle. It contains a total of 21,500 RS images collected by an unmanned aerial vehicle from a height of 100 meters. This dataset consists of 16136 and 5364 cactus and non-cactus images, respectively. All images of the cactus and non-cactus classes are of the same size (32×32). Figure 2 presents examples of the columnar cactus plant images. The considered dataset consists of two image folders: train and test. We randomly split the train folder, which consists of 17500 images, into 80% and 20% for training and validation, respectively. Once the model completed training, the test folder, which consists of 4000 images is employed to evaluate the performance. To get better results, we employed different data augmentation techniques to generate more images for training [27]. Data augmentation is employed to enlarge the training dataset, while the valid and test data are not augmented. Scaling, horizontal flip, random rotation (10 degrees), zoom, intensity shift, and lighting conditions are the augmentation strategies applied to the training dataset.

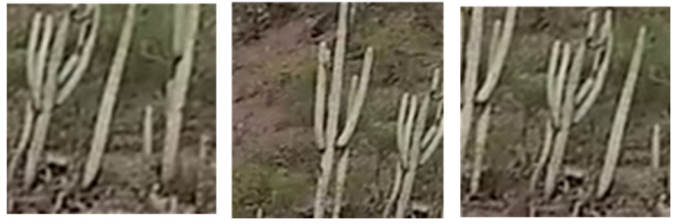

Fig. 2. Examples of columnar cactus plant images from the considered dataset

*3.3 Experimental Results*

Precision, recall, accuracy, loss, and f1-score measures are used to evaluate the proposed model. The overall performance of the ERI-CNN was tested using 4000 images. As shown in Table 2, in terms of accuracy, the proposed network yields 99% in the training phase, 97% in the validation phase, and 98% in the testing phase.

Table 2: ERI-CNN accuracy results

| Proposed model | Training accuracy | Validation accuracy | Test accuracy |
|---|---|---|---|
| ERI-CNN | 99 % | 97% | 98% |

Besides, the proposed model achieves 97% of precision, 97% of recall, and 97.5% of f1-score, as shown in Table 3. The obtained loss value is very low (0.056), which indicates that only a few errors are made by the model.

Table 3: ERI-CNN performance

| Accuracy | Precision | Recall | F1-score | Loss |
|---|---|---|---|---|
| 98% | 97% | 97% | 97.5% | 0.056 |

*3.4 Impact of Data Augmentation*

To reveal the impact of data augmentation on the obtained results, we trained the ERI-CNN without using the augmentation techniques. The classification accuracy was dropped from 98% to 96%. It is noticeable that the loss value is high compared to the first obtained loss, which indicates that the errors made by the data classification are more significant. The results depicted in Table 4 demonstrate the impact of the data augmentation in improving the proposed ERI-CNN performance.

Table 4: The impact of data augmentation on the ERI-CNN performance

| Model | Accuracy | Precision | Recall | F1-score | Loss |
| --- | --- | --- | --- | --- | --- |
| Without data augmentation | 96% | 93.5% | 97% | 95% | 0.241 |
| With data augmentation | 98% | 97% | 97% | 97.5% | 0.056 |

## 4. Comparison with Other Networks

To validate the performance of the proposed ERI-CNN, we compare the obtained results with other models. The used cactus aerial photos dataset was employed in [19]. In this work, the authors have proposed an approach based on a modified LeNet-5 CNN and they obtained an accuracy of 95%. Compared to the obtained results in [19], we have achieved a significant accuracy improvement of 3%.

Furthermore, using transfer learning, we fine-tuned the InceptionV3 model [28] and compared the achieved results. We loaded a pre-trained version of the model. The accuracy of the InceptionV3 model was 96%, while we obtained 98%. This is demonstrated in Fig. 3., which illustrates the two confusion matrices resulted from the ERI-CNN and the InceptionV3 model using the same test dataset.

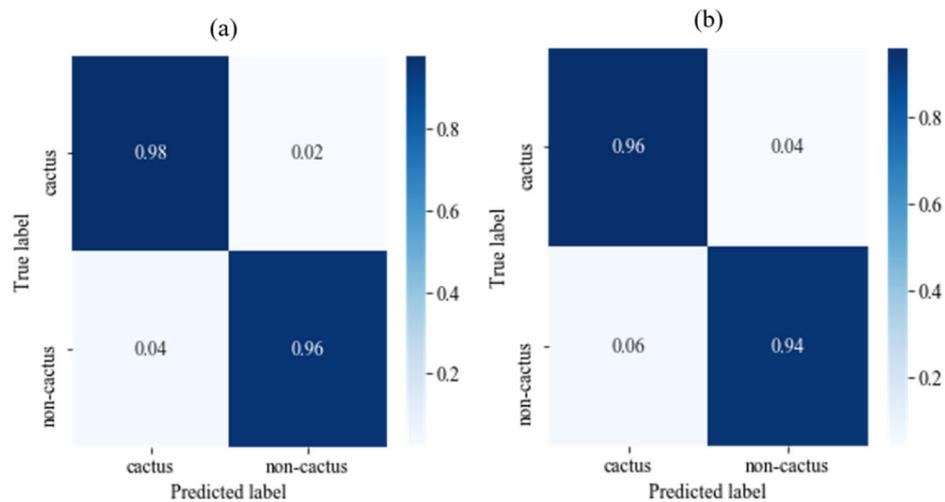

Fig. 3. (a) Confusion matrix of the ERI-CNN (b) Confusion matrix of the InceptionV3

From Fig. 4., we depict that our ERI-CNN achieves a good precision, recall, and f1-score for both cactus and non-cactus classes compared to the InceptionV3 results.

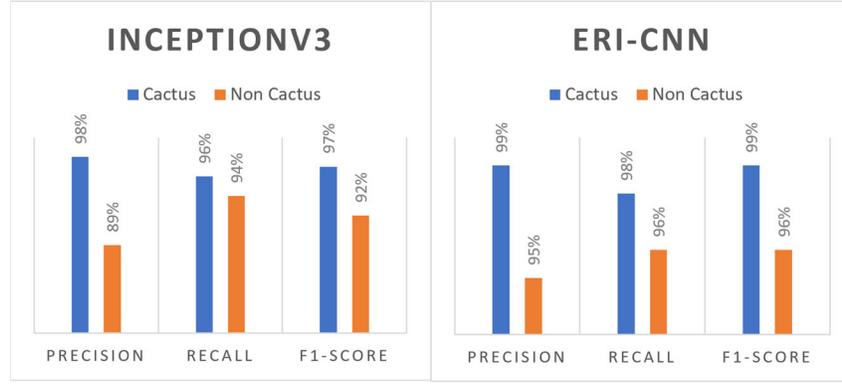

Fig. 4. Precision, recall, and f1-score values of InceptionV3 and our ERI-CNN

Table 5 illustrates a comparative study between 3 different DL models, which are: our proposed ERI-CNN, InceptionV3, and a modified LetNet-5 CNN [19].

Table 5: The obtained accuracy using different types of initialization and different DL models.

| Parameters | ERI-CNN | InceptionV3 | A modified LeNet-5 CNN [19] |
| --- | --- | --- | --- |
| Initializer | Proposed initialization method | ImageNet pre-trained weights | Not indicated |
| Number of epochs | 100 | 100 | 150 |
| Training time | 51mn | 120mn | Not indicated |
| Test accuracy | 98% | 96% | 95% |

This study shows that our proposed ERI-CNN model has a better performance with a test accuracy of 98% compared to InceptionV3 and the modified LeNet-5 CNN. This indicates that our proposed model with its deep layers could extract several features, although with random weights.

In fact, the used initialization method, the network structure, the configuration parameters, and the type of layers are the main factors in achieving a higher recognition rate compared to other networks. The used initialization method starts by choosing the slope angles of neurons in random from a target interval, then randomly rotates the activation functions around the y-axis, and ends by assigning weights in harmony with the distribution of input data. Using this method gives better results than other initialization methods, where the slope of the activation functions is controlled, therefore the degree of the model generalization.

Besides, the ERI-CNN model combines nine convolutional layers that have different dimensions of filter size of (2×2) and (3×3). Consequently, employing different filters size for the convolutional layers permits learning and covering more various features, therefore improving the classification results.

## 5. Conclusion

In this work, we opted to develop and test a randomly initialized CNN for columnar cactus recognition. We used a huge public dataset with more than 21000 RS images. Insightful results with an accuracy of 98%, a precision of 97%, and a recall of 97% were obtained, which demonstrate the high performance of the proposed model. This model can be adapted to perform automatic surveillance for protected natural areas. Our future goals will be to extend the model with a wider variety of training datasets and to explore other current research trends such as tree counting and tree species classification to support smart forestry and sustainable agricultural practices [29][30].

# References

[1] M. Stoyanova, Y. Nikoloudakis, S. Panagiotakis, E. Pallis, and E. K. Markakis. (2020) "A survey on the internet of things (IoT) forensics: challenges, approaches, and open issues." *IEEE Communications Surveys and Tutorials* **22 (2):** 1191-1221.

[2] H. N. Dai, H. Wang, G. Xu, J. Wan, and M. Imran. (2020). "Big data analytics for manufacturing internet of things: opportunities, challenges and enabling technologies." *Enterprise Information Systems* **14 (9-10):** 1279-1303.

[3] T. Meng, X. Jing, Z. Yan, and W. Pedrycz. (2020). "A survey on machine learning for data fusion." *Information Fusion* **57:** 115-129.

[4] C. Tian, L. Fei, W. Zheng, Y. Xu, W. Zuo, and C. W. Lin. (2020) "Deep learning on image denoising: An overview." *Neural Networks* **131:** 251-275.

[5] S. Ben Atitallah, M. Driss, W. Boulila, and H. Ben Ghézala. (2020) "Leveraging Deep Learning and IoT big data analytics to support the smart cities development: Review and future directions." *Computer Science Review* **38:** 100303.

[6] Y. Hajjaji, W. Boulila, I. R. Farah, I. Romdhani, and A. Hussain, (2021) "Big data and IoT-based applications in smart environments: A systematic review." *Computer Science Review* **39:** 100318.

[7] W. Boulila, M. Sellami, M. Driss, M. Al-Sarem, M. Safai, and F. A. Ghaleb. (2021) "RS-DCNN: A Novel Distributed Convolutional-Neural-Networks based-Approach for Big Remote-Sensing Image Classification." *Computers and Electronics in Agriculture* **182:** 106014.

[8] M. Mohammadi, A. Al-Fuqaha, S. Sorour, and M. Guizani. (2018) "Deep learning for IoT big data and streaming analytics: A survey." *IEEE Communications Surveys & Tutorials* **20 (4):** 2923-2960.

[9] I. Chebbi, W. Boulila, N. Mellouli, M. Lamolle, and I. R. Farah. (2018) "A comparison of big remote sensing data processing with Hadoop MapReduce and Spark." In *4th International Conference on Advanced Technologies for Signal and Image Processing (ATSIP)*, IEEE, 1-4.

[10] W. Boulila, Z. Ayadi, and I. R. Farah. (2017) "Sensitivity analysis approach to model epistemic and aleatory imperfection: Application to Land Cover Change prediction model." *Journal of computational science* **23:** 58–70.

[11] A. Ferchichi, W. Boulila, and I. R. Farah. (2017) "Towards an uncertainty reduction framework for land-cover change prediction using possibility theory." *Vietnam Journal of Computer Science* **4 (3):** 195–209.

[12] L. Ma, Y. Liu, X. Zhang, Y. Ye, G. Yin, and B. A. Johnson. (2019) "Deep learning in remote sensing applications: A meta-analysis and review." *Journal of Photogrammetry and Remote Sensing* **152**: 166–177.

[13] M. Chi, A. Plaza, J.A. Benediktsson, Z. Sun, J. Shen, and Y. Zhu. (2016) "Big data for remote sensing: Challenges and opportunities." *Proceedings of the IEEE* **104 (11):** 2207–2219.

[14] D. Hong, L. Gao, N. Yokoya, J. Yao, J. Chanussot, Q. Du, and B. Zhang. (2020) "More diverse means better: Multimodal deep learning meets remote-sensing imagery classification." *IEEE Trans. on Geoscience and Remote Sensing* **59(5):** 4340 – 4354.

[15] Q. Yuan, H. Shen, T. Li, Z. Li, S. Li, Y. Jiang, H. Xu, W. Tan, Q. Yang, J. Wang, and J. Gao. (2020) "Deep learning in environmental remote sensing: Achievements and challenges." *Remote Sensing of Environment* **241:** 111716.

[16] W. Li, H. Fu, L. Yu, and A. Cracknell. (2017) "Deep learning based oil palm tree detection and counting for high-resolution remote sensing images." *Remote Sensing* **9 (1):** 22.

[17] T. Kattenborn, J. Eichel, S. Wiser, L. Burrows, F.E Fassnacht, and S. Schmidtlein. (2020) "Convolutional Neural Networks accurately predict cover fractions of plant species and communities in Unmanned Aerial Vehicle imagery." *Remote Sensing in Ecology and Conservation*.

[18] Y. Zhu, W. Sun, X. Cao, C. Wang, D. Wu, Y. Yang, and N. Ye. (2019) "TA-CNN: Two-way attention models in deep convolutional neural network for plant recognition." *Neurocomputing* **365:**191–200.

[19] E. López-Jiménez, J. I. Vasquez-Gomez, M. A. Sanchez-Acevedo, J. C. Herrera-Lozada, and A. V. Uriarte-Arcia. (2019) "Columnar cactus recognition in aerial images using a deep learning approach." *Ecological Informatics* **52:** 131–138.

[20] Z. Zheng, S. Pan, and Y. Zhang. (2019) "Fruit Tree Disease Recognition Based on Convolutional Neural Networks." In *IEEE International Conferences on Ubiquitous Computing & Communications and Data Science and Computational Intelligence and Smart Computing, Networking and Services:* 118–122.

[21] S. Nezami, E. Khoramshahi, O. Nevalainen, I. Pölönen, and E. Honkavaara. (2020) "Tree species classification of drone hyperspectral and rgb imagery with deep learning convolutional neural network." *Remote Sensing* **12 (7):** 1070.

[22] G. Hu, Y. Zhu, M. Wan, W. Bao, Y. Zhang, D. Liang, and C. Yin, "Detection of diseased pine trees in unmanned aerial vehicle images by using deep convolutional neural networks." *Geocarto International*: 1–20.

[23] C. Gallicchio and S. Scardapane. (2020) "Deep Randomized Neural Networks." In *Recent Trends in Learning from Data* **896**: 43–68.

[24] H. Li, M. Krček, and G. Perin. (2020) "A Comparison of Weight Initializers in Deep Learning-based Side-channel Analysis." In *International Conference on Applied Cryptography and Network Security*: 126–143.

[25] G. Dudek. (2019) "Improving randomized learning of feedforward neural networks by appropriate generation of random parameters." In *International Work-Conference on Artificial Neural Networks*: 517–530.

[26] D. P. Kingma and J. L. Ba. (2015) "Adam: A method for stochastic optimization." In *3rd International Conference on Learning Representations*.

[27] C. Shorten and T. M. Khoshgoftaar. (2019) "A survey on Image Data Augmentation for Deep Learning." *Journal of Big Data* **6 (1)**.

[28] C. Szegedy, W. Liu, Y. Jia, P. Sermanet, S. Reed, D. Anguelov, D. Erhan, V. Vanhoucke, and A. Rabinovich. (2015) "Going deeper with convolutions." In *Proceedings of the IEEE conference on computer vision and pattern recognition*: 1–9.

[29] W. Zou, W. Jing, G. Chen, Y. Lu, and H. Song. (2019) "A survey of big data analytics for smart forestry." *IEEE Access* **7:** 46621-46636.

[30] J. Delgado, N. M. Short, D. P. Roberts, and B. Vandenberg. (2019) "Big Data Analysis for Sustainable Agriculture." *Frontiers in Sustainable Food Systems* **3:**54.